\documentclass{article}
\usepackage{arxiv}
\usepackage{graphicx,epsfig}
\usepackage{dcolumn}
\usepackage{xcolor}
\usepackage{lineno}
\usepackage{amsfonts}
\usepackage{amssymb}
\usepackage{mathtools}
\usepackage{bm}
\usepackage{qcircuit}
\usepackage{braket}
\usepackage{float}
\usepackage{algorithm}
\usepackage{algorithmicx}
\usepackage{algpseudocode}
\usepackage{pdfpages}
\usepackage{times}

\usepackage{soul}
\usepackage{url}
\usepackage{booktabs}
\usepackage{amsmath}
\usepackage[normalem]{ulem}
\usepackage{diagbox}

\newcommand{\bea}{\begin{equation}}
\newcommand{\eea}{\end{equation}\noi}
\newcommand{\ber}{\begin{eqnarray}}
\newcommand{\eer}{\end{eqnarray}\noi}
\title{Financial Vision Based Differential Privacy Applications}

\author{{\hspace{1mm}Jun-Hao Chen} \\
	Department of Computer Science and\\
	Information Engineering\\
	National Taiwan University\\
	\And
	{\hspace{1mm}Yi-Jen Wang}\\
	Department of Data Science\\
	Soochow University\\
	\AND
	{\hspace{1mm}Yun-Cheng Tsai}\\
	Department of Data Science\\
	Soochow University\\
	\texttt{pecutsai@gm.scu.edu.tw}
	\And
	{\hspace{1mm}Samuel Yen-Chi Chen}\\
	Computational Science Initiative\\
	Brookhaven National Laboratory
}

\begin{document}
\maketitle

\begin{abstract}
The importance of deep learning data privacy has gained significant attention in recent years. It is probably to suffer data breaches when applying deep learning to cryptocurrency that lacks supervision of financial regulatory agencies. However, there is little relative research in the financial area to our best knowledge. We apply two representative deep learning privacy-privacy frameworks proposed by Google to financial trading data. We designed the experiments with several different parameters suggested from the original studies. In addition, we refer the degree of privacy to Google and Apple companies to estimate the results more reasonably. The results show that DP-SGD performs better than the PATE framework in financial trading data. The tradeoff between privacy and accuracy is low in DP-SGD. The degree of privacy also is in line with the actual case. Therefore, we can obtain a strong privacy guarantee with precision to avoid potential financial loss.
\end{abstract}

\section{\label{sec:intro}Introduction}
The Digital Currency market size has unprecedentedly exceeded two trillion in May 2021. The market has three times more growth than the remarkable peak value in Dec 2017. The trading volume of the total cryptocurrency market also surpasses a hundred billion marks (Coinmarketcap.com accessed on Aug 1st, 2021). According to PwC's survey~\cite{pwc2021digital} in 2020, the total assets under management (AuM) of crypto hedge funds increased from \$2 to \$3.8 billion. Moreover, the median AuM at fund launch is \$1 million, a 15 times increase than the previous year. These noticeably growing phenomena show that the cryptocurrency market has reached a high activity and maturity in these years. Besides, many financial services and trading activities in the fiat currency market have headed into the cryptocurrency market.

With the rapid evolution and outstanding development of artificial intelligence (AI), many innovative financial technology (FinTech) applications applied to cryptocurrency. For instance, AI-based quantitative investment algorithm~\cite{alessandretti2018anticipating, jiang2017cryptocurrency, xie2020blockchain}, anti-money laundering detection~\cite{weber2019anti}, decentralized federated learning framework~\cite{bonawitz2019towards, li2020blockchain}, and more. Although these deep learning applications have already successfully provided financial services, these deep learning algorithms are still vulnerable to privacy breaches. Privacy protection is a critical issue for companies with massively private and sensitive client data, especially in cryptocurrency that lacks adequate supervision by financial regulatory agencies. The importance of data privacy has gained significant attention in recent years. The European Union (EU) specifies a data privacy and security law to protect personal information rights. General Data Protection Regulation (GDPR) puts an effect on May 25, 2018.
Moreover, the governor of California also approved California Consumer Privacy Act (CCPA) in June 2018. It became effective in 2020; It gives consumers more control over their personal information. Therefore, building the AI-based FinTech service with privacy-preserving is the key to reliable FinTech services in the next generation.

Training a high-accuracy neural network without sacrificing data privacy is the end goal of the AI system. A reasonable way to achieve this aim is to apply security techniques of the information domain to neural networks. Differential privacy is one of the practical solutions to provide a strict privacy guarantee to sensitive data~\cite{park2019attack, zhang2020broadening}. Various research on applying differential privacy to neural networks has been proposed in recent years. PATE and DP-SGD are two representative approaches proposed by Google to offer the privacy guarantee in deep learning. The following are the brief introductions for two methods:

\begin{itemize}
    \item \textbf{Private Aggregation of Teacher Ensembles (PATE):} 
    
    PATE is an ensemble model-free approach that can provide a privacy guarantee by coordinating the inference of different machine learning models. PATE preserves privacy with a specific framework. Still, the whole procedure is complex and has more assumptions.

    \item \textbf{Differentially Private Stochastic Gradient Descent (DP-SGD):} 

    DP-SGD achieves private directly by modifying the classic Stochastic gradient descent (SGD) algorithm with differential privacy. DP-SGD has fewer assumptions than PATE but is much complex in the training algorithm.
\end{itemize}

This study designs an experiment to examine the practical results of applying the privacy-preserving training mechanism to the actual cryptocurrency data. We first emulate a trading patterns recognition task as a simple AI-based fintech application. Then based on this simple deep learning task, we apply both PATE and DP-SGD methods and measure the privacy results with Renyi Differential Privacy (RDP). With the RDP measurement, we can observe the training result more objectively. Moreover, we also compare our results to other data types in the practical industries. In the end, we will discuss the benefits and drawbacks of adopting the deep learning privacy-preserving mechanism to cryptocurrency data.

The remainder of this study is arranged as follows: Section 2 gives an overview of candlestick patterns' backgrounds to design the simple pattern recognition task. Section 3 provides the preliminary of differential privacy and relative algorithms. Section 4 presents our experimental design and settings, and Section 5 shows the results and discussions. At last, Section 6 gives the conclusion of our study and our contribution.

\section{\label{sec:bg}Backgrounds}
\subsection{\label{subsec:candlestick}Candlestick Chart}
Price data is the most common data type in finance. It is a sequence of deal prices decided by the free market mechanism.  These prices are also called \emph{quoted prices} reported from the financial exchange center. However, it is challenging to inspect helpful information directly from the raw price data. Until the 17th century, a Japanese designed a data representation method called \emph{candlestick chart} to observe rice's price changes better~\cite{nison2001japanese}. Candlestick chart represents a series of raw price data into four representative prices: open, high, low, and close. Generally, we call them \emph{OHLC} in short. Based on the OHLC, many analysis tools like technical analysis and morphology are proposed.

\begin{figure}[htbp]
\centering
\includegraphics[scale=0.5]{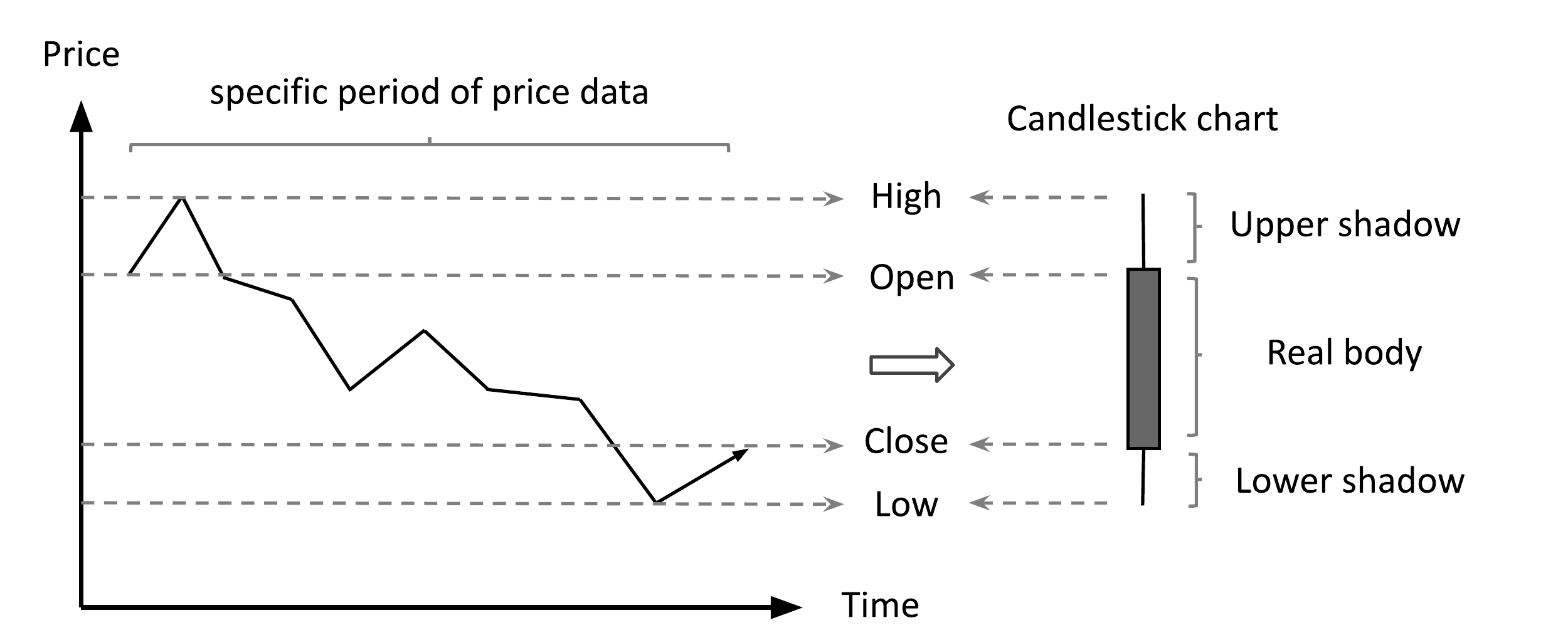}
\caption{The left-hand side shows the raw price data, and the right-hand side shows the candlestick represented by the left-hand side. A black candlestick here means a dropping candle.}
\label{candlestick_intro}
\end{figure}

\subsection{\label{subsec:candlestick_pattern}Candlestick Patterns}
The prediction of these trading prices has been one of the most critical and challenging issues for a long time. Not only is the amount of financial data very massive, but the relation between the data is nonlinear and full of noise~\cite{sharma2017survey}. Therefore, it is challenging to build a stable predictive model with the unprocessed pricing data~\cite{ratto2018ensemble}. However, with the human traders' domain, \emph{Candlestick Chart} is a standard used intuitive visualized method for extracting refined features from raw pricing data. Traders and investors have already used it to observe price variation and seek potentially profitable opportunities for decades. Many types of research showed many possibilities to speculate through candlestick patterns on the different targets in the financial market.

\subsection{\label{subsec:gafcnn}GAF-CNN Model}
GAF-CNN model consists of two phases:
\begin{enumerate}
    \item time-series encoding process by using the Gramian Angular Field (GAF) \cite{wang2015encoding} method,
    \item train the Convolutional Neural Networks (CNN) model with the encoded matrice.
\end{enumerate}
We elaborate on the two phases in the following. Phase 1: Time-series encoding process. Based on the matrice type inputs format, it is necessary to encode the time-series data as images for the model. There are three steps in this phase. Firstly, we rescale the input time-series data
$X={{x}_{1}, {x}_{2},\ldots,{x}_{n}}$
using the minimum-maximum scaling in Equation \ref{equ:minmax} to make every $\widetilde{x}_{i}$ to the interval $[0, 1]$. The notation $\widetilde{x}_i$ indicates each normalized time-series belongs to the entire normalized set $\widetilde{X}$. Secondly, adopting Equation \ref{equ:gaf_arccos} to transform $\widetilde X$ into the polar coordinate system. Thirdly, we then create the GAF matrix with the angles as Equation \ref{equ:gaf}. The timeline of the GAF matrix starts from the top left to the bottom right.   

\begin{align}
\widetilde{x}_{i}&=\frac{x_i-\min(X)}{\max(X)-\min(X)}
\label{equ:minmax}
\end{align}

\begin{equation}
\begin{aligned}
\phi &= \arccos(\widetilde{x}_i), 0\leq \widetilde{x}_i \leq 1, \widetilde{x}_i \in \widetilde{X}\\
r &= \frac{t_i}{N}, t_i\in\mathbb{N}
\label{equ:gaf_arccos}
\end{aligned}
\end{equation}

\begin{equation}
\begin{aligned}
\textup{GAF} &=\cos(\phi_i + \phi_j) \\ &=
\left [\begin{matrix}
\cos(\phi_{1}+\phi_{1}) & \cdots & \cos(\phi_{1}+\phi_{n}) \\
\cos(\phi_{2}+\phi_{1}) & \cdots & \cos(\phi_{2}+\phi_{n}) \\
\vdots & \ddots & \vdots \\
\cos(\phi_{n}+\phi_{1}) & \cdots & \cos(\phi_{1}+\phi_{n})
\end{matrix}\right]
\label{equ:gaf}
\end{aligned}
\end{equation}

Phase 2 is to input the encoded matrice to the CNN model. We will receive a symmetric $n$ by $n$ time-series GAF matrix after phase 1, where $n$ denotes the length of time-series data. The GAF matrix is full of temporal information of time-series data. The diagonal and non-diagonal elements in the GAF matrix contain the original and correlation of every two points of the time-series data. Additionally, it is bijective after normalizing the GAF matrix to $[0, 1]$. Therefore, the GAF matrix can be transformed to the normalized time-series data with the diagonal elements. Compared to the abstract GAF matrix, it is a more useful and intuitive way to understand the results of experiments.

\section{\label{sec:meth}Methodology}
\subsection{\label{subsec:dp}Differential Privacy (DP)}
Deep learning algorithms have a wide variety of applications and have been highly developed in recent years. The issue of model security has gradually received attention, especially the sensitive training data such as personal privacy, medical diagnosis, confidential information of the company, and more. Without any protection, deep learning models are vulnerable to the negative impact of attacks when deploying to the actual product~\cite{akhtar2018threat, fredrikson2015model, chakraborty2018adversarial}. Differential privacy~\cite{dwork2006our} is one of the effective methods to protect models from specific attacks by adding noise to sensitive data. It constitutes a robust privacy guarantee due to its composability, robustness to auxiliary information, and graceful degradation in the presence of correlated data~\cite{abadi2016deep}.

Differential privacy was introduced by Dwork et al. in 2006~\cite{dwork2006our}. Differential privacy preserves privacy by adding random sampled noise to the original data. The noise can be sampled from different distributions. The Laplace and Gaussian noise perform better at numeric data, and the exponential distribution is suitable for non-numeric data. Therefore, the measurement of privacy costs is also a critical stage. It helps confirm the robustness of privacy guarantees for algorithms on current data. Equation~\ref{equ:dp} provides $(\varepsilon, \delta)$ - differential privacy for any two neighbor inputs $d$ and $d'$ $\in$ $D$~\cite{dwork2016concentrated}. Notation $M$ is a random mechanism. In the case of deep learning, $M$ usually denotes the model itself. Privacy budget $\epsilon$ quantifies the primary privacy risk. A smaller $\epsilon$ represents preserving more privacy. Parameter $\delta$ is defined as the possibility that allows for privacy to be broken. A larger $\delta$ extends the limitation of the total differential privacy~\cite{zhu2017preliminary}.
\begin{equation}
\begin{aligned}
Pr[M(d) \in \mathbb{R}] \le e^{\epsilon}\cdot Pr[M(d') \in \mathbb{R}] + \delta
\end{aligned}
\label{equ:dp}
\end{equation}

The following Section~\ref{subsec:rdp} presents an advanced way to compute privacy costs considered one of the commonly used privacy standards in recent years. Section~\ref{subsec:pate} and Section~\ref{subsec:pate} respectively illustrate two typical frameworks to achieve the same goal of privacy-preserving machine learning. Most of the notations used in this section referred to the original research.

\subsection{\label{subsec:rdp}Renyi Differential Privacy (RDP)}
Renyi Differential Privacy (RDP) can track privacy costs more directly and share several important characteristics compared with differential privacy. It also provides stronger privacy guarantees than DP~\cite{ dwork2016concentrated, bun2016concentrated}. 
Equation~\ref{equ:rdp} is the definition of RDP. The complete mathematical derivation can be found in~\cite{mironov2017renyi}. ($\alpha$, $\epsilon$)-RDP and $\epsilon$-DP are equal when $\alpha$ = $\infty$. RDP will satisfied $(\epsilon - \frac{\log\delta}{\alpha - 1}, \delta)$-DP, therefore it provides stronger privacy guarantees than DP for any 0 $< \delta <$ 1.

\begin{equation}
\begin{aligned}
d_\alpha(M(d)\|M(d')) \le \epsilon
\end{aligned}
\label{equ:rdp}
\end{equation}

\subsection{\label{subsec:pate}Private Aggregation of Teacher Ensembles (PATE)}
\begin{figure}[htbp]
\centering
\includegraphics[scale=0.35]{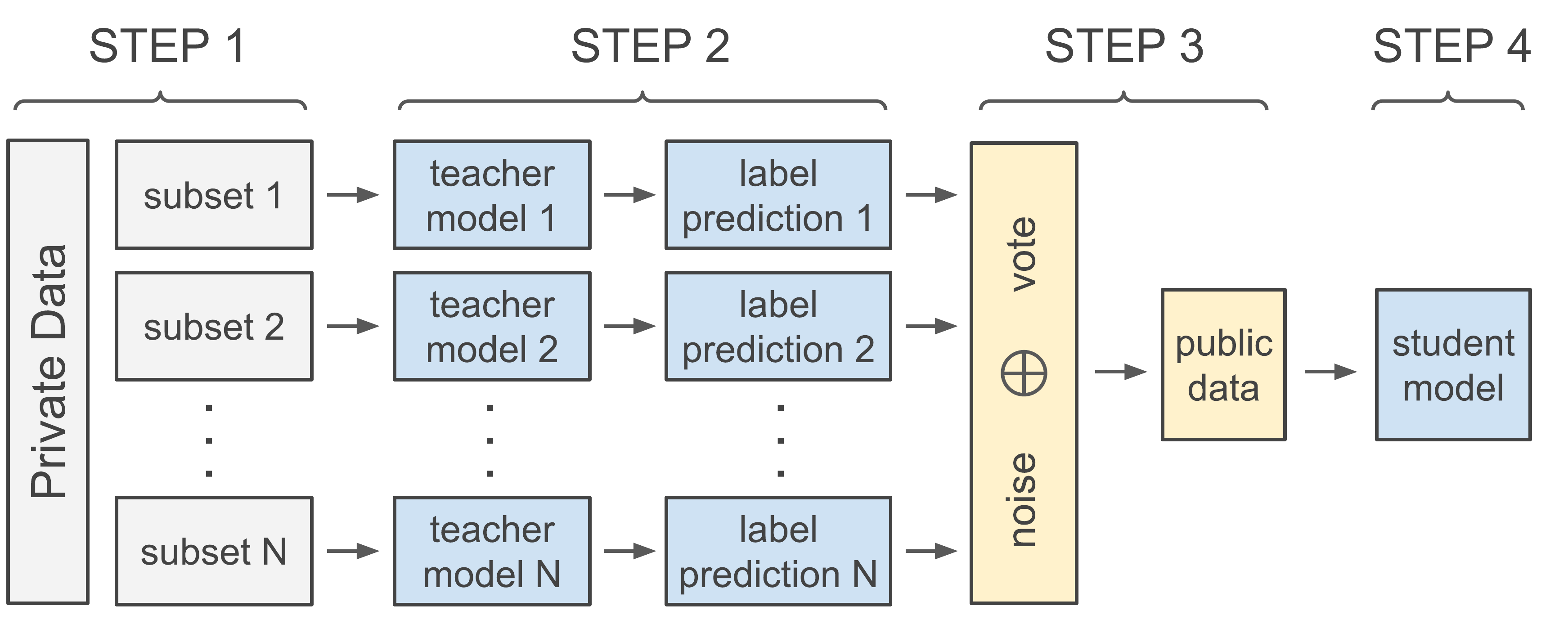}
\caption{The flow diagram of the Private Aggregation of Teacher Ensembles (PATE) framework.}
\label{fig:pate_flowchart}
\end{figure}

Private Aggregation of Teacher Ensembles (PATE)~\cite{papernot2016semi} is a traditional way to preserve privacy in machine learning. PATE protects privacy with a specific framework of knowledge aggregation. PATE contains four steps, and Figure~\ref{fig:pate_flowchart} shows the complete flowchart.
\begin{itemize}
    \item \textbf{Step 1. Partition private data:} 
    Partition the private data into N disjoint subsets. Each disjoint subset will use to train the individual teacher model. This ensures that no pair of teacher models will be trained on overlapping data.
    \item \textbf{Step 2. Train and predict teacher models and predict:} 
    After training with disjoint data subsets, use teacher models to predict public datasets labels. There is no limit on teacher models, and any model or even different models can use in this step.
    \item \textbf{Step 3. Label aggregation:}
    Count the number of teacher models voted for each class, then add random noise sampled from the specific distribution to the vote counts.  The distribution adopted here can refer to the differential privacy research in Section 3. Adding noise from specific distribution in this step satisfies privacy guarantees of differential privacy.
    \item \textbf{Step 4. Training student model:}
    A student model is trained with public data, labeled by the aggregate output of the ensemble of teacher models. The student model will learn to imitate ensemble teacher models without private data. It is impossible to peep into private data. Therefore, even if someone successfully attacks the student model with implicit memorization, they can only recover the public data.
\end{itemize}

For example, in the case of 10 teacher models solving a simple binary classification task. The vote counts of class 1, maybe 4.

\subsection{\label{subsec:dpsgd}Differentially Private Stochastic Gradient Descent (DP-SGD)}
Abadi et al.~\cite{abadi2016deep} propose the Differentially Private Stochastic Gradient Descent (DP-SGD) algorithm in 2016 to introduce differential privacy into the model training algorithm. DP-SGD is more general than PATE but more modifications to the training mechanism with fewer assumptions. DP-SGD is literally a modification of the minibatch SGD algorithm to add privacy into models' parameters. Algorithm~\ref{alg:dpsgd} shows the process in detail. Loading training data, setting hyperparameters, and initializing models' weights are the same as the classic SGD algorithm. The main differences are in the model updating process:
\begin{enumerate}
    \item \textbf{Compute and clip gradient:} 
    DP-SGD clips the gradient with gradient norm bound $C$ after computing $g_{t}(x_{i})$. Clipping gradient helps control the influence that each training sample can impact model parameters. This ensures that a single training sample will not affect the overall update too much.
    \item \textbf{Add random noise:}
    The random Gaussian noise added to the sum up clipped gradients depends on the noise scale $\sigma$. This noise multiplier control how much noise is added to the gradients. In general, more noise added can give the model more privacy protection.
    \item \textbf{Gradient descent:}
    Make the gradient descent as usual.
    \item \textbf{Compute privacy cost:}
    RDP, which was mentioned before, tracks privacy loss. Privacy loss is changed with random noise added and will gradually accumulate during the training process. At the end of the training, we will compute an $\epsilon$ value to show the sum of privacy losses in the entire training process. According to the size of the value to define, is the privacy guarantee meaningful.
    \end{enumerate}

\begin{algorithm}[tb]
\begin{algorithmic}
\State \textbf{Load} training data $X = \{x_{1}, x_{2}, ..., x_{N}\}$
\State \textbf{Set} loss function $\mathcal{L}(\theta) = \frac{1}{N} \sum_{i=1}^{N}\mathcal{L}(\theta, x_{i})$.
\State \textbf{Set} learning rate $\eta_{t}$, noise scale $\sigma$, group size $L$, and gradient norm bound $C$.
\State \textbf{Initialize} $\theta_{0}$ randomly.
\For{\texttt{$t$ in $T$}}
    \State Random sample a $L_{t}$ via sampling probability $\frac{L}{N}$.
    \State \textbf{Step 1. Compute and clip gradient}
    \For{\texttt{$i$ in $L_{t}$}}
        \State compute $g_{t}(x_{i}) \leftarrow \nabla_{\theta_{t}} \mathcal{L}(\theta_{t}, x_{i})$
        \State $\bar{g}_{t}(x_{i}) \leftarrow g_{t}(x_{i})\;/\;max(1, \frac{\|g_{t}(x_{i})\|_{2}}{C})$
    \EndFor
    \State \textbf{Step 2. Add random noise}
    \State $\tilde{g}_{t} \leftarrow \frac{1}{L}(\sum_{i=1}^{L}\bar{g}_{t}(x_{i}) + \mathcal{N}(0, \sigma^2C^2I))$
    \State \textbf{Step 3. Gradient descent}
    \State $\theta_{t+1} \leftarrow \theta_{t} - \eta_{t}\cdot\tilde{g}_{t}$
\EndFor
\State \textbf{Step 4. Compute privacy cost}
\State \textbf{Output} $\theta_{T}$ and compute the privacy cost $(\varepsilon, \delta)$ with a privacy accounting method.
\end{algorithmic}
\caption{Differentially Private Stochastic Gradient Descent (DP-SGD)}
\label{alg:dpsgd}
\end{algorithm}

\section{\label{sec:exp}Experiment}
\subsection{\label{subsec:data}Data Illustration}
Although the candlestick chart is of great use for observing the variation of prices, it is still challenging to extract useful information to determine the arbitrage opportunities in the financial market. With the long-term accumulated experience in practice, traders show that specific \emph{candlestick patterns} provide valuable information about the dynamic financial market. These candlestick patterns can capture the subtle variety of supply and demand behind the financial markets. However, obtaining excellent candlestick patterns from professional traders can be very expensive. To avoid assigning expert labeling, we refer to \emph{The Major Candlesticks Signals}~\cite{stephen2014major} and handmade eight candlestick patterns with the critical rules.

\begin{figure}[htbp]
\centering
\includegraphics[width=0.65\textwidth]{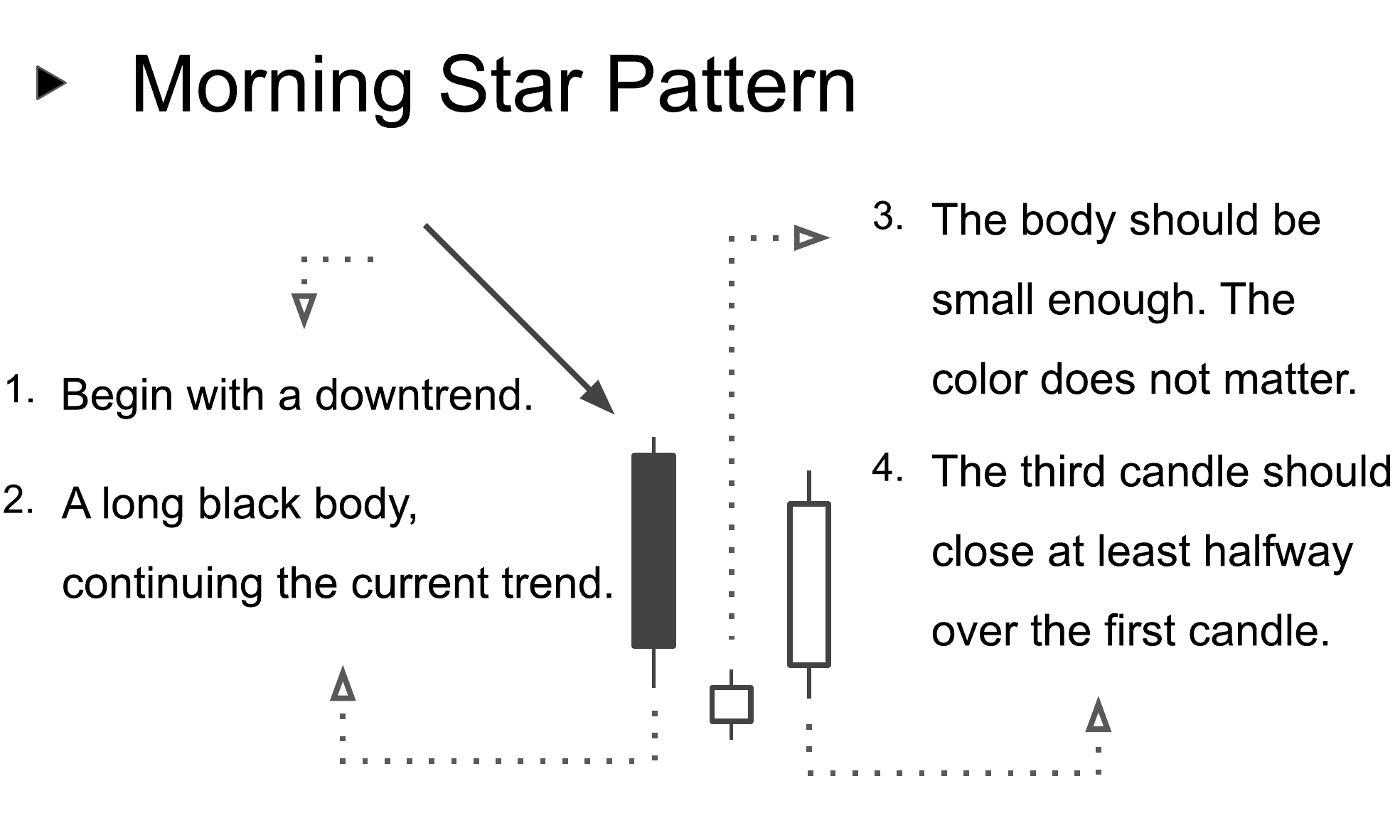}
\caption{{\bfseries Illustration of the Morning Star Pattern.} The candlesticks present the shape of the Morning Star pattern. The texts display the major rules depicted in the book.}
\label{morning_intro}
\end{figure}

The candlestick patterns adopted in our handmade data include the Morning Star, Evening Star, Bullish Engulfing, Bearish Engulfing, Shooting Star, Inverted Hammer, Bullish Harami, and Bearish Harami patterns. Figure~\ref{morning_intro} is an example to illustrate the Morning Star Pattern. Morning Star is a bottom reversal signal that begins with a downtrend; the first candle should be a long black body, continuing the current trend. The second candle is a small enough body, and the third candle should close at least half over the first candle. This pattern considers as a bottom reversal signal.

The ETH/USD 1-minute OHLC data from January 1, 2010, to January 1, 2021, is used to handmade these eight candlestick patterns. 

\subsection{\label{subsec:expdesign}Experimental Design}
DP technique has been well-developed in these years. Many representative international companies, such as Google and Apple, apply DP to protect clients' privacy in various products and services. The noise sizes setting in different industries are very different. There is no specific standard for noise setting so far. Moreover, the DP technique is less mentioned in finance compared to the technology industry. It makes applying privacy-preserving models to finance much more tricky. 

This study aims to compare the differences in differential privacy between financial data and the data in the literature. We first design a simple candlestick pattern recognition model and tune the optimal hyperparameter. Next, two privacy-preserving deep learning mechanisms, PATE and DP-SGD, are applied to the original model. We will analyze the result with MNIST and CIFAR-10 datasets. Moreover, we will discuss the optimal noise size in our financial experiments to the values used in the actual company.

\subsection{\label{subsec:expsetup}Experimental Setup}
\begin{figure}[htbp]
\centering
\includegraphics[scale=0.35]{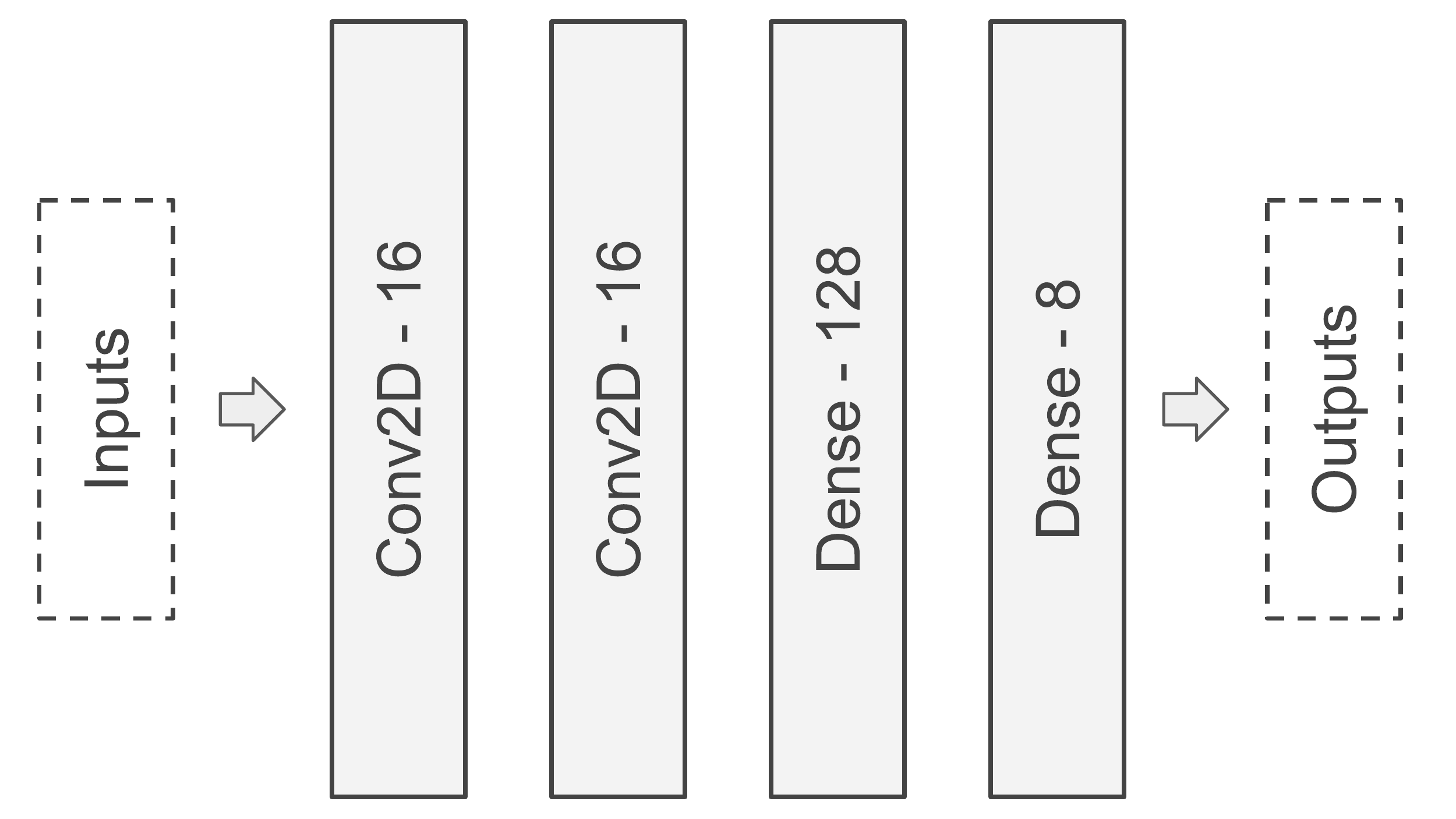}
\caption{This is the simple architecture of the baseline CNN model. The pooling layers are not suitable for encoded time-series data~\cite{chen2020encoding}. Therefore, pooling layers are not adopted in this study.}
\label{fig:cnn_arch}
\end{figure}

We create a simple CNN model for the baseline model with two convolutional layers followed by one fully connected layer. We train the model with the python package OPTUNA~\cite{Optuna} from 100 trials to automatic search optimal hyperparameters. The standard SGD with momentum is used for the optimizer. The detail parameters are listed in Table~\ref{tab:best hyperparameters}, and the architecture of the baseline CNN model shows in Figure~\ref{fig:cnn_arch}. In PATE's training procedure, we split the training data into 10, 20, and 50 disjoint subsets and then train teacher models on these disjoint subsets, respectively. In PATE's training procedure, we split the training data into 10, 20, and 50 disjoint subsets and then train teacher models on these disjoint subsets, respectively. The noise is added before training the student model. It is added with the Laplacian mechanism on the aggregate output of the ensemble of teachers before training the student model. The noise of inversed scale $\gamma$ between 0.01 and 1. based on the~\cite{papernot2016semi}. In the DP-SGD case, we directly adopt the DP-SGD optimizer published by Google TensorFlow privacy package~\cite{DP_SGD}. Referring to suggestions of the previous, we set the parameters as follows: the gradient clip parameter to 1 and 1.5; the noise scale to between 0.01 and 1~\cite{abadi2016deep}.

\begin{table}[htbp]
\centering
\setlength{\arrayrulewidth}{0.3mm}
\renewcommand\arraystretch{1.5}
\begin{center}
\begin{tabular}{c|c} 
\hline
hyperparameter & value\\ [0.5ex] 
\hline
Learning rate & 0.006\\ 
Momentum & 0.9 \\
Batch size & 100 \\
PATE epochs & 100 \\
DP-SGD epochs & 120 \\
\hline
\end{tabular}
\end{center}
\smallskip
\caption{The best baseline model hyperparameters.}
\label{tab:best hyperparameters}
\end{table}

\section{\label{sec:result}Result and Discussion}
\begin{figure}[htbp]
\centering
\includegraphics[scale=0.5]{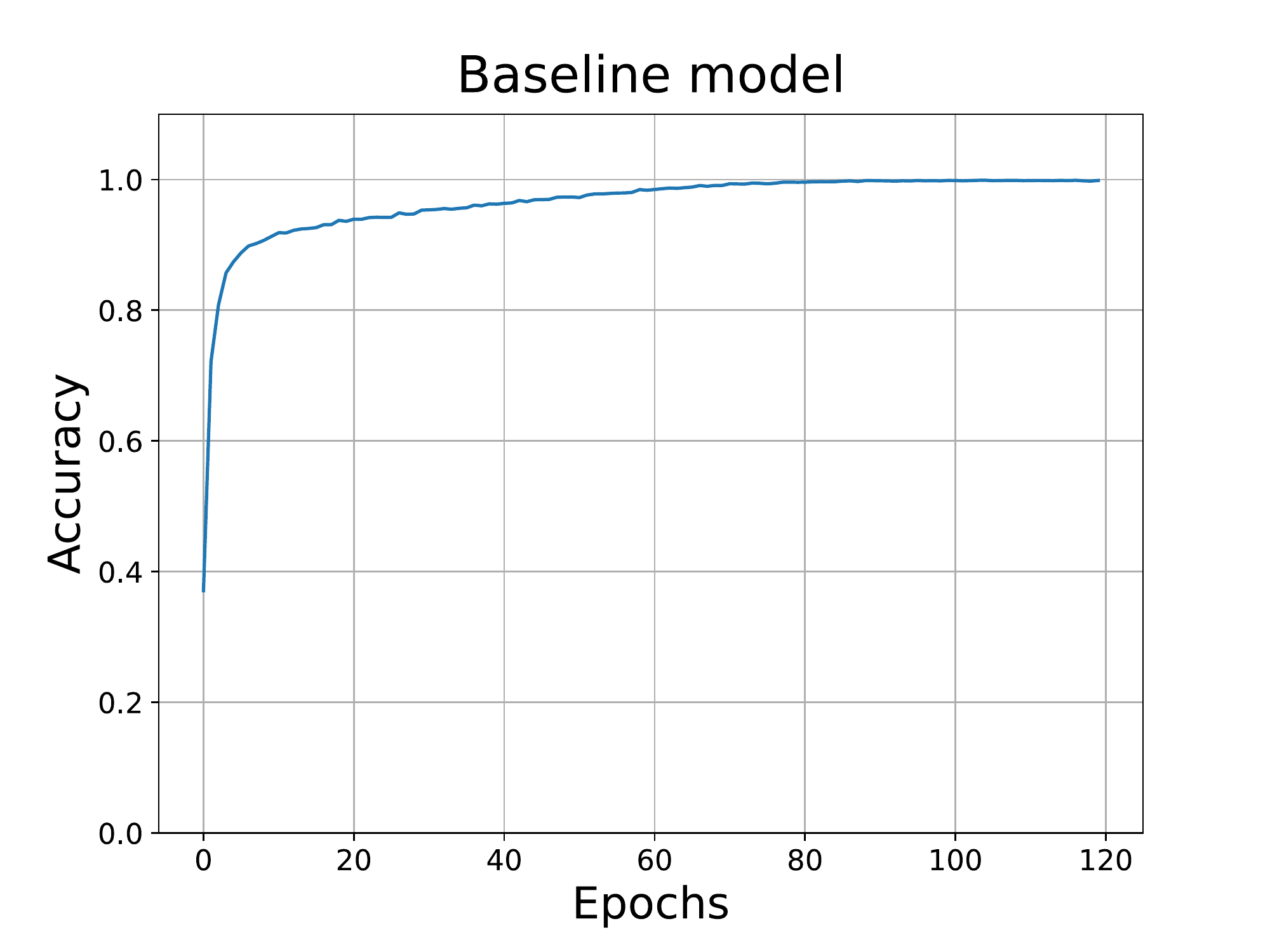}
\caption{Baseline model training process on accuracy}
\label{fig:baseline}
\end{figure}

The best accuracy of the baseline model under 100 trails can achieve 93.17\%. The training process is shown in Figure~\ref{fig:baseline}. We will compare the differences between PATE and DP-SGD methods in our financial dataset below.

\subsection{\label{subsec:result_measurement}Privacy Measurement}
Before discussing the results, we need to know under which noise value ($\epsilon$) represents a privacy guarantee model. The original research recommends using $\epsilon$ less than 1 for privacy protection~\cite{dwork2014algorithmic}. However, it will cause significant difficulties in data interpretation in actual cases~\cite{garfinkel2019understanding,mervis2020researchers}. Several representative companies choose to use higher $\epsilon$ to balance the privacy protection and data readability tradeoff. It loses parts of privacy protection but keeps parts of the data readability. Google uses $\epsilon=9$ in their services. Apple uses $\epsilon=6$ in MacOS and $\epsilon=14$ in the IOS system~\cite{greenberg2017one}. Therefore, we choose to use $\epsilon=14$ as the standard of comparison. If the model still has enough accuracy with $\epsilon<=14$, the result is practical useable.

\subsection{\label{subsec:result_pate}PATE results}
\begin{figure}[htbp]
\centering
\includegraphics[scale=0.5]{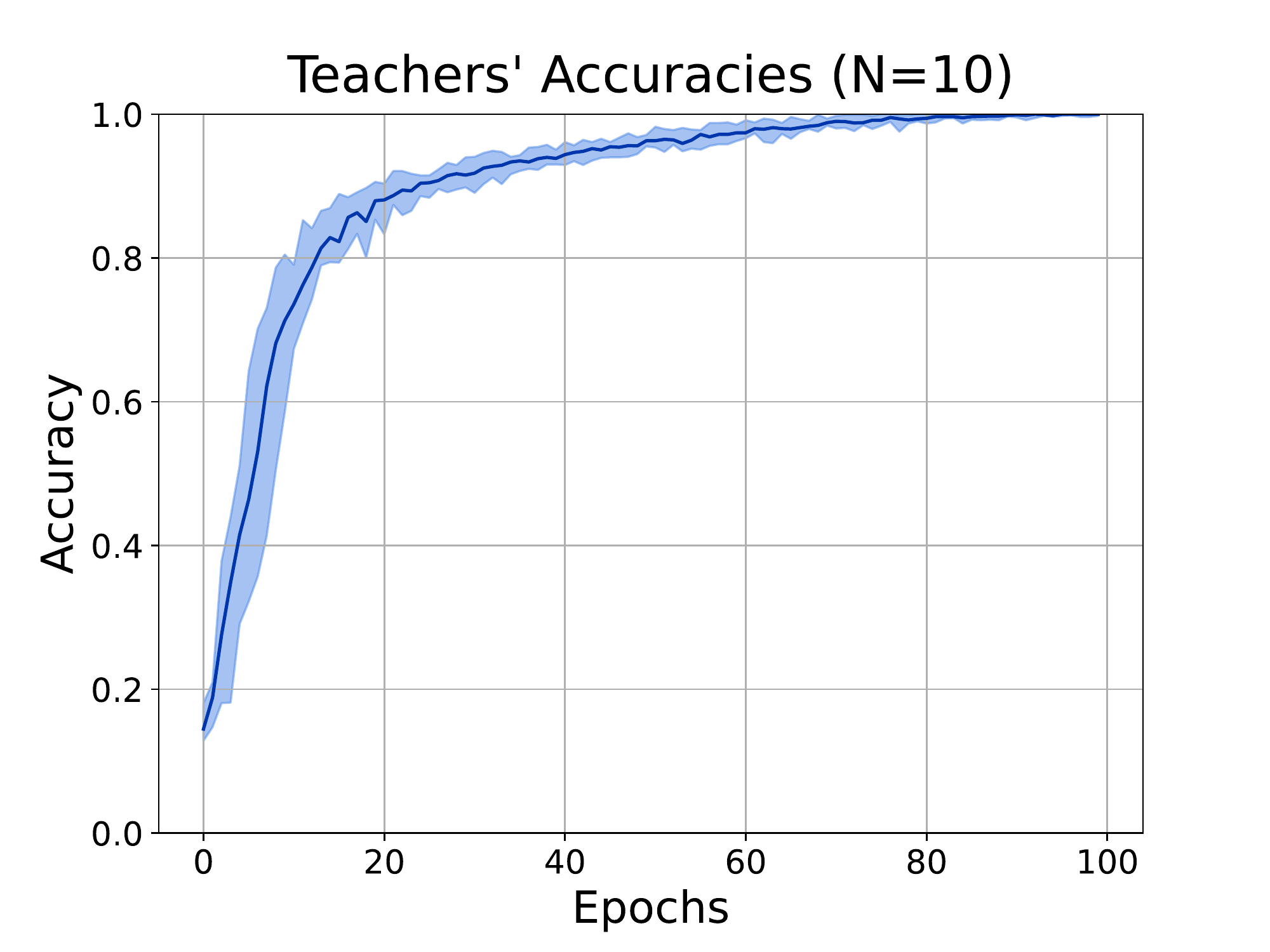}
\caption{Teacher models training process with N=10}
\label{fig:pate_tm_n10}
\end{figure}

\begin{figure}[htbp]
\centering
\includegraphics[scale=0.5]{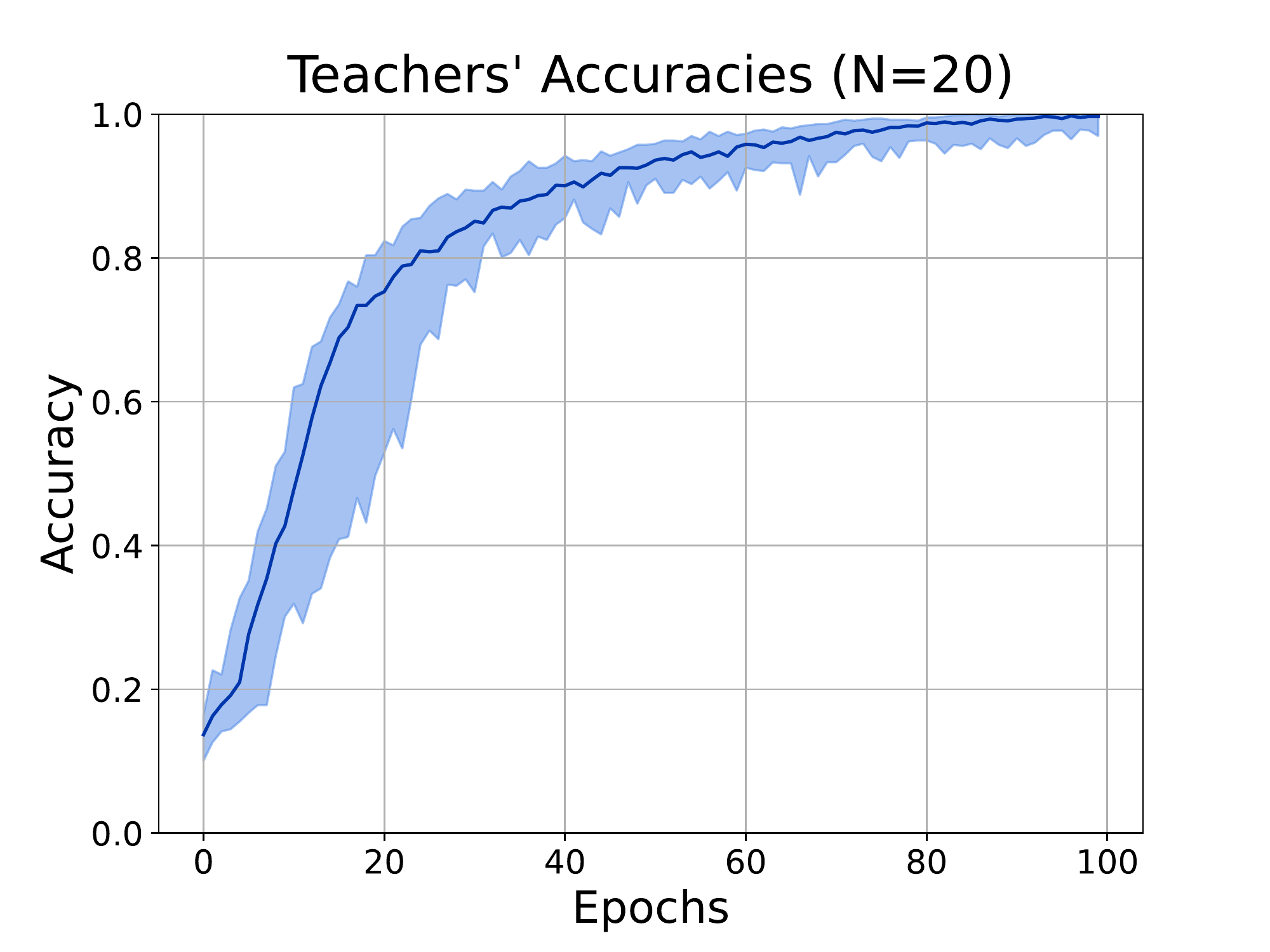}
\caption{Teacher models training process with N=20}
\label{fig:pate_tm_n20}
\end{figure}

\begin{figure}[htbp]
\centering
\includegraphics[scale=0.5]{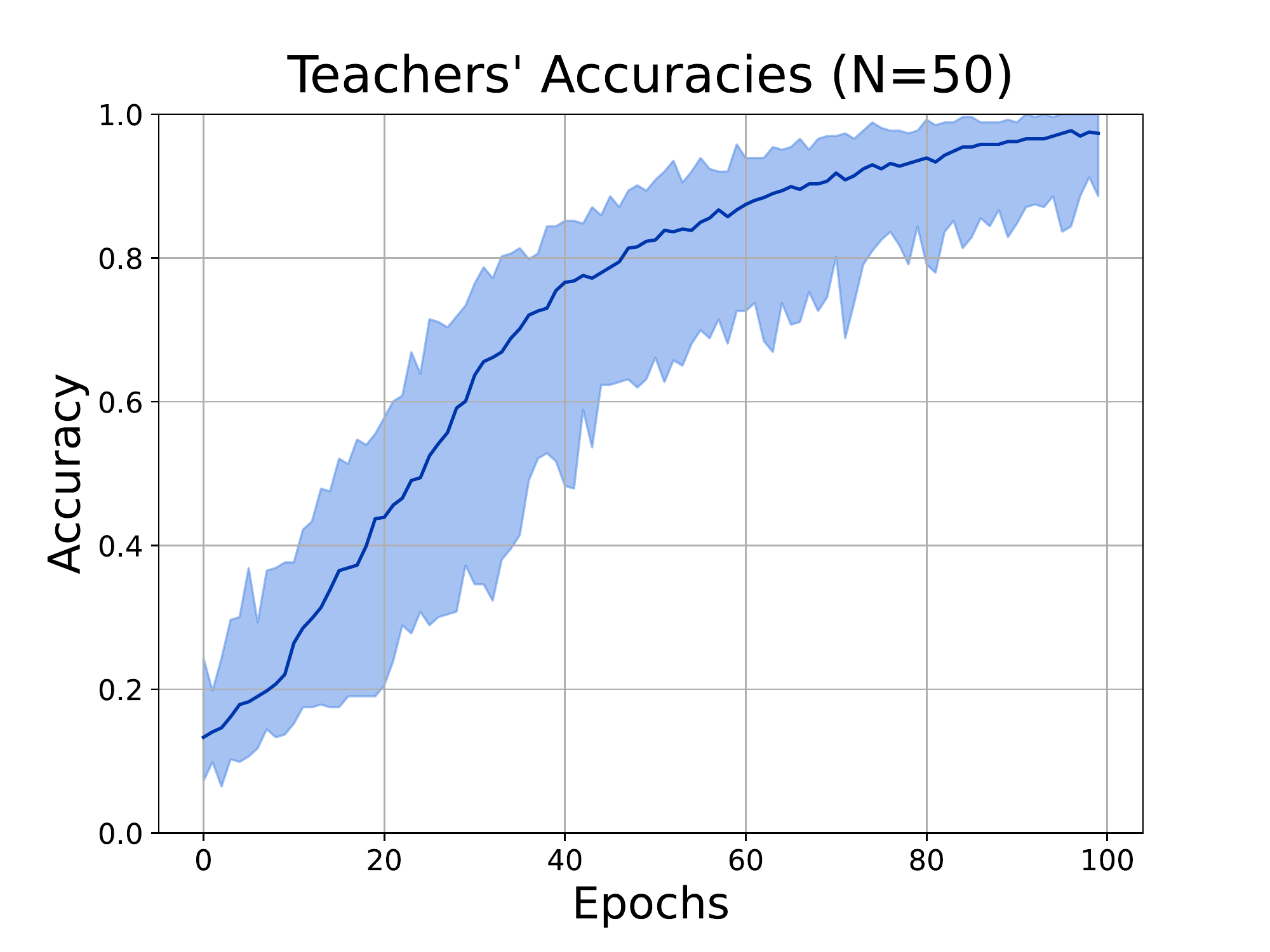}
\caption{Teacher models training process with N=50}
\label{fig:pate_tm_n50}
\end{figure}

The training processes on teacher models with N=10, 20, and 50 are shown in Figures~\ref{fig:pate_tm_n10},~\ref{fig:pate_tm_n20} and~\ref{fig:pate_tm_n50}, respectively. The blue area in the figures is the accuracy range of N teacher models, and the middle line shows the median value of the accuracies. As all figures show, we have two main results in PATE experiments. First, the smaller the N is, the more stable the accuracy of the student model is. The accuracy is high and converges to epoch 100 in the training process. However, the stability among teacher models with N=10, 20, and 50 is significantly different. N=10 shows a perfect converge process, but the training process becomes unstable as the number increases like N=50. This result is reasonable and matches the previous literature due to fewer training samples as N increases.

Second, the smaller the $\epsilon$ is, the less accurate the student model is. The best accuracy of the student model can reach 91.92\%, which is close to the non-private baseline model of 93.17\%, but it almost has no privacy protection. However, when the $\epsilon$ comes to 16.28, the best accuracy can only reach 60.54\%. The phenomenon shows that there is a tradeoff between accuracy and privacy-preserving. This result makes sense according to the literature. Moreover, we can find that the models are almost impracticable when the $\epsilon$ is lower than 14. It is tough to balance accuracy and privacy-preserving in the PATE framework, even in such a simple CNN architecture.

\begin{table}[htbp]
\centering
\setlength{\arrayrulewidth}{0.3mm}
\renewcommand\arraystretch{1.5}
\begin{tabular}{c|ccc}
\hline
\diagbox{Noise($\epsilon$)}{Teacher}&N=10&N=20&N=50\\
\hline
1(53781.36) & 91.92\% & 91.13\% & 89.05\%\\
10(636.94) & 76.35\% & 85.92\% & 87.48\%\\
30(95.25) & 58.80\% & 66.80\% & 80.96\%\\
50(43.10) & 49.76\% & 61.02\% & 72.71\%\\
100(16.28) & 49.07\% & 53.24\% & 60.54\%\\
\hline
\end{tabular}
\smallskip
\caption{Student test accuracies in different teacher N and noise.}
\label{tab:PATE student accuracy}
\end{table}

\subsection{\label{subsec:resilt_dpsgd}DP-SGD results}
\begin{figure}[htbp]
\centering
\includegraphics[scale=0.5]{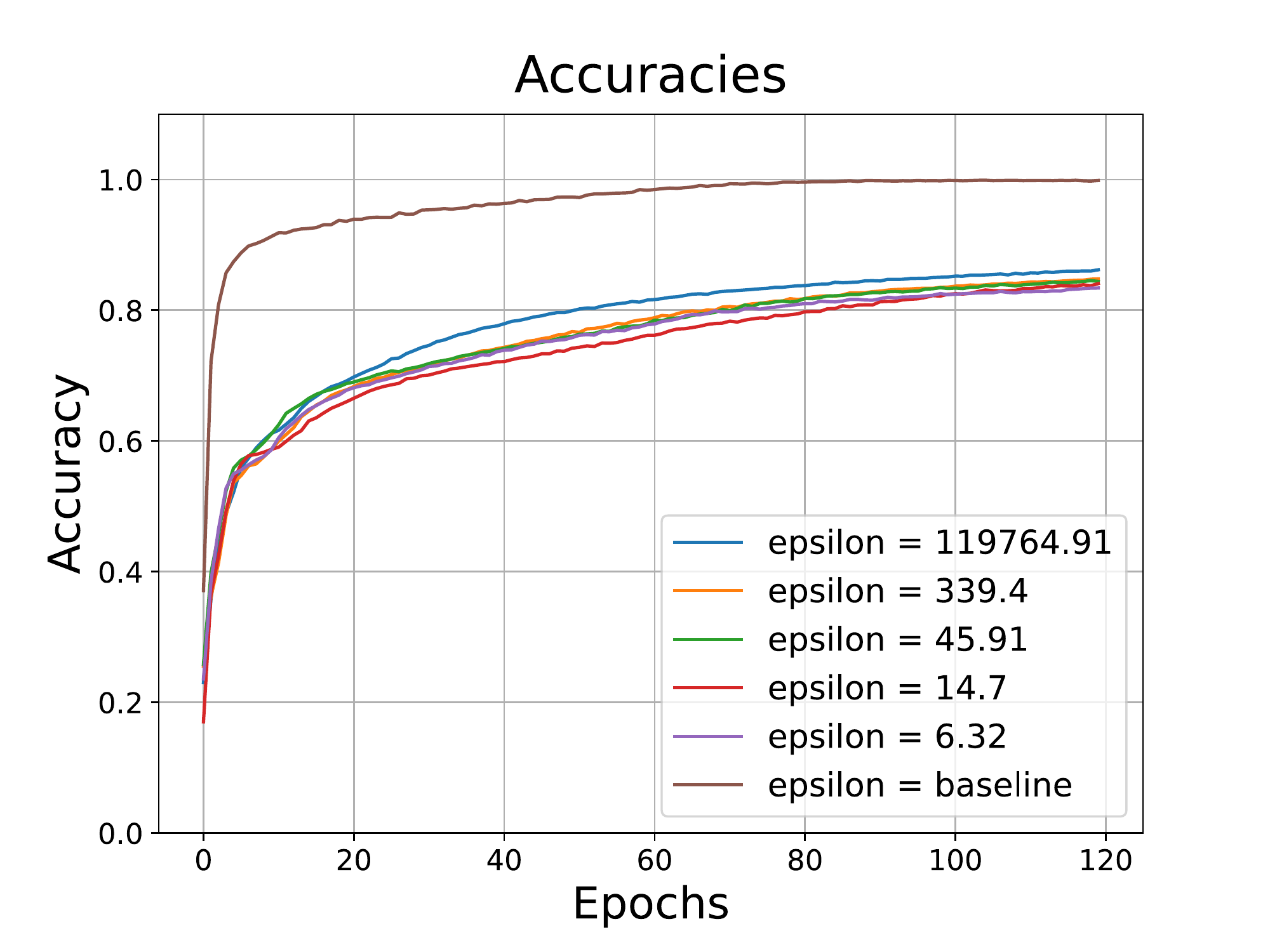}
\caption{Results on accuracy for different noise levels when gradient clipping equals 1.0}
\label{fig:dpsgd_clip_1}
\end{figure}

\begin{figure}[htbp]
\centering
\includegraphics[scale=0.5]{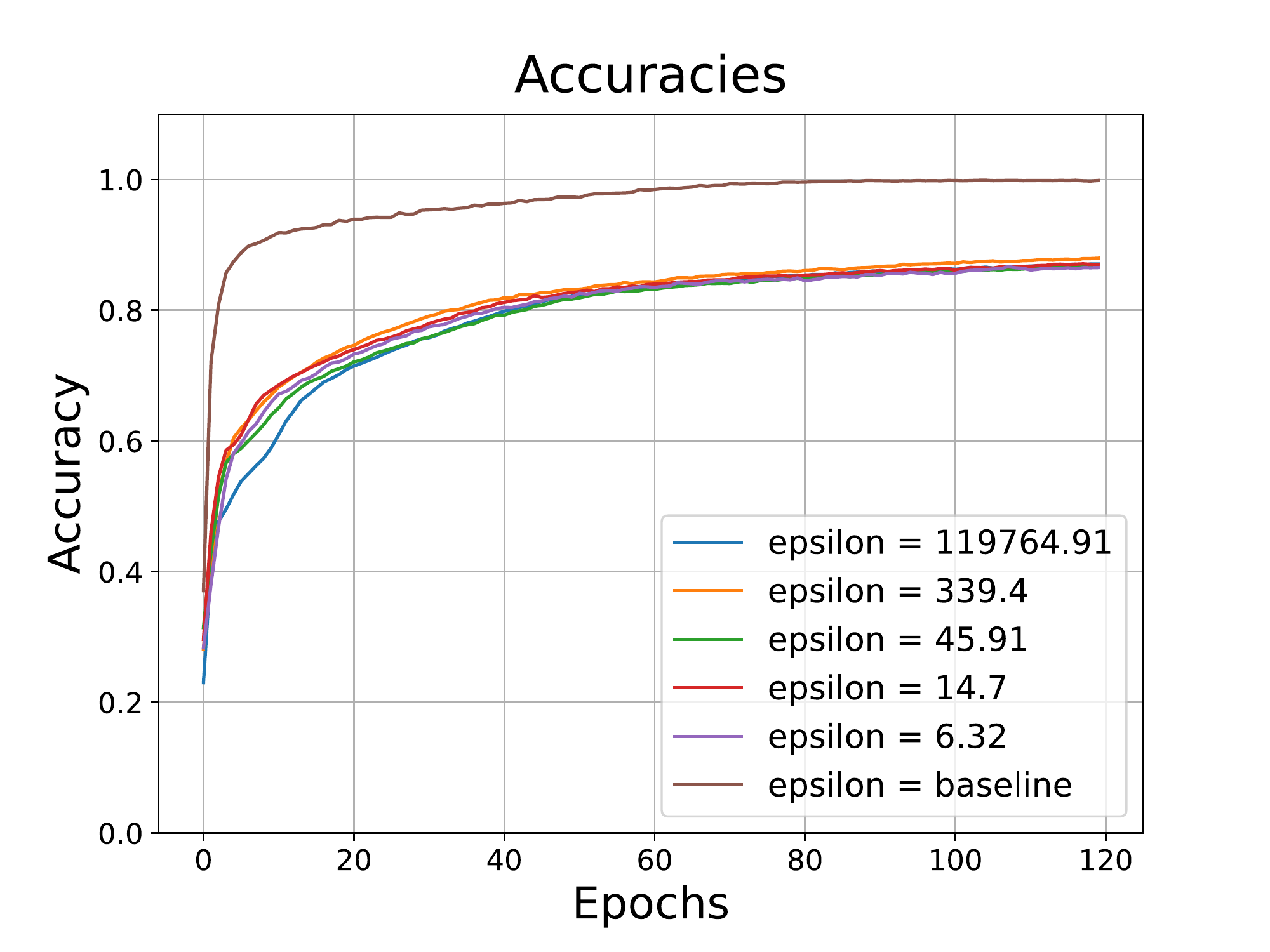}
\caption{Results on accuracy for different noise levels when gradient clipping equals 1.5}
\label{fig:dpsgd_clip_1.5}
\end{figure}

According to the original study, gradient norm clipping size is suggested to be set between 1 and 1.5. Therefore, we did two experiments with clipping sizes equal to 1 and 1.5 to see the difference. Figures~\ref{fig:dpsgd_clip_1} and~\ref{fig:dpsgd_clip_1.5} show the results of clipping sizes equals to 1 and 1.5, respectively. The results show that the accuracy is more stable when the gradient clipping is set to 1.5 than 1. In Figure10, even the lowest $\epsilon$ has similar accuracy and stable coverage process as the largest $\epsilon$. This means that the accuracy and the privacy-preserving cost are low. 

Table~\ref{tab:DP-SGD accuracy} shows the best training results from different $\epsilon$ and gradient clipping sizes. All experiment results perform significantly better than PATE cases when $\epsilon$ is lower than 14. The model can still reach an 87.98\% value when $\epsilon$ is low to 6.32, significantly different from the PATE framework. It is noteworthy that the noise-adding mechanism is other between PATE and DP-SGD frameworks. Therefore, the noise scale adopted in both experiment frameworks has a significant difference. Both noise scales adopted in this study are recommended from the original researches.

\begin{table}[htbp]
\centering
\setlength{\arrayrulewidth}{0.3mm}
\renewcommand\arraystretch{1.5}
\begin{tabular}{c|cc}
\hline
\diagbox{Noise($\epsilon$)}{Clipping} & 1 & 1.5\\
\hline
0.1(119764.91) & 88.18\% & 89.48\%\\
0.3(339.40) & 86.92\% & 88.53\\
0.5(45.90) & 86.31\% & 88.39\%\\
0.7(14.70) & 85.79\% & 88.17\% \\
1(6.32) & 85.27\% & 87.98\% \\

\hline
\end{tabular}
\smallskip
\caption{Test accuracies in different gradient clipping and noise.}
\label{tab:DP-SGD accuracy}
\end{table}

DP-SGD method is also more directly to realize the privacy-preserving training. Overall, the results of DP-SGD perform much better than the PATE framework at accuracy and stability. DP-SGD can remain almost 88\% accuracy with $\epsilon$ is low to 6.32. This result is very close to the non-private baseline model accuracy. However, PATE's models are impracticable when $\epsilon$ is lower than 100, with no privacy guarantee.

\section{\label{sec:conclusion}Conclusion}
Financial sectors possess lots of sensitive personal data. They will suffer enormous losses with data breaches, especially in cryptocurrency that lacks adequate supervision by financial regulatory agencies. Therefore, model security is essential to help financial technology reach the next milestone. In this study, we apply two famous privacy-preserving frameworks proposed by Google to financial data. To have a reasonable comparison standard, we refer to Google and Apple and use $\epsilon=14$ to estimate the privacy protection. If the model still has enough accuracy with $\epsilon<=14$, then the model has an adequate privacy guarantee. We present several parameter setting results in both frameworks, which are not mentioned in financial data. Our results show that DP-SGD performs better than PATE in almost all cases. DP-SGD is also more directly to be used. The model can still perform well in our financial data when the $\epsilon$ value is low to the value of 6.2 in DP-SGD but is almost undertrained in PATE. Therefore, training a financial model with a robust privacy guarantee is feasible.

\bibliographystyle{bmc-mathphys}
\bibliography{./bib/cryptocurrency,
              ./bib/classical_ml,
              ./bib/finance,
              ./bib/attack,
              ./bib/privacy,
              ./bib/blockchain_ai}
\end{document}